\documentclass[11pt,a4paper]{article}
\usepackage{times,latexsym}
\usepackage{url}
\usepackage[T1]{fontenc}

\usepackage[acceptedWithA]{tacl2018}

\usepackage{xspace,mfirstuc,tabulary}

\newif\iftaclinstructions
\taclinstructionsfalse 
\iftaclinstructions
\renewcommand{\confidential}{}
\renewcommand{\anonsubtext}{(No author info supplied here, for consistency with
TACL-submission anonymization requirements)}
\newcommand{\instr}
\fi

\iftaclpubformat 

\else

\fi

\usepackage{booktabs}
\usepackage{tabularx}
\usepackage{todonotes}
\usepackage{multirow}
\usepackage{url}
\usepackage{amsmath}
\usepackage{amssymb}
\usepackage{tablefootnote}
\usepackage{caption}
\usepackage{subcaption}
\usepackage{bbm}

\newcommand{\regdsl}[1]{\texttt{\small #1}}

\newcommand{\prog}{r}
\newcommand{\hole}{\square}

\newcommand{\sketch}{{{S}}}
\newcommand{\lang}{{{L}}}

\newcommand{\hh}{{\hat{h}}}
\newcommand{\bh}{{\bar{h}}}

\newcommand{\hw}{{\hat{w}}}

\DeclareMathOperator*{\argmax}{arg\,max}

\newcommand{\figref}[1]{Figure~\ref{fig:#1}}
\newcommand{\figlabel}[1]{\label{fig:#1}}
\newcommand{\deepregex}{\textsc{DeepRegex}}
\newcommand{\drex}{\textsc{DeepRegex+Filter}}
\newcommand{\deepsketch}{\textsc{DeepSketch}}
\newcommand{\gramsketch}{\textsc{GrammarSketch}}
\newcommand{\semregex}{{\sc SemRegex}}
\newcommand{\sempre}{{\sc Sempre}}
\newcommand{\kb}{\textsc{KB13}}
\newcommand{\turk}{\textsc{Turk}}
\newcommand{\so}{\textsc{StackOverflow}}

\title{Sketch-Driven Regular Expression Generation \\from Natural Language and Examples}

\author{
 Xi Ye$^\diamondsuit$ \quad  Qiaochu Chen$^\diamondsuit$  \quad  Xinyu Wang$^\spadesuit$ \quad Isil Dillig$^\diamondsuit$ \ \ \ \ \ Greg Durrett$^\diamondsuit$ \\
 $^\diamondsuit$ Department of Computer Science, The University of Texas at Austin \\
 $^\spadesuit$ Computer Science and Engineering Department, University of Michigan, Ann Arbor \\
  $^\diamondsuit${\texttt{\{xiye,qchen,isil,gdurrett\}@cs.utexas.edu} } \\
  $^\spadesuit${\texttt {xwangsd@umich.edu}}
}

\date{}

\begin{document}
\maketitle
\begin{abstract}
Recent systems for converting natural language descriptions into regular expressions (regexes) have achieved some success, but typically deal with short, formulaic text and can only produce simple regexes. Real-world regexes are complex, hard to describe with brief sentences, and sometimes require examples to fully convey the user's intent. 
We present a framework for regex synthesis in this setting where both natural language (NL) and examples are available.
First, a semantic parser (either grammar-based or neural) maps the natural language description into an intermediate \emph{sketch}, which is an incomplete regex containing holes to denote missing components. Then a program synthesizer searches over the regex space defined by the sketch and finds a regex that is consistent with the given string examples. Our semantic parser can be trained purely from weak supervision based on correctness of the synthesized regex, or it can leverage heuristically-derived sketches.
We evaluate on two prior datasets \cite{KB13,deepregex} and a real-world dataset from Stack Overflow. Our system achieves state-of-the-art performance on the prior datasets and solves 57\% of the real-world dataset, which existing neural systems completely fail on.\footnote{Code and data available at \url{https://github.com/xiye17/SketchRegex/}}  
\end{abstract}

\section{Introduction}

\begin{figure*}[ht]
\small
\includegraphics[width=\textwidth,trim=130 105 130 100,clip]{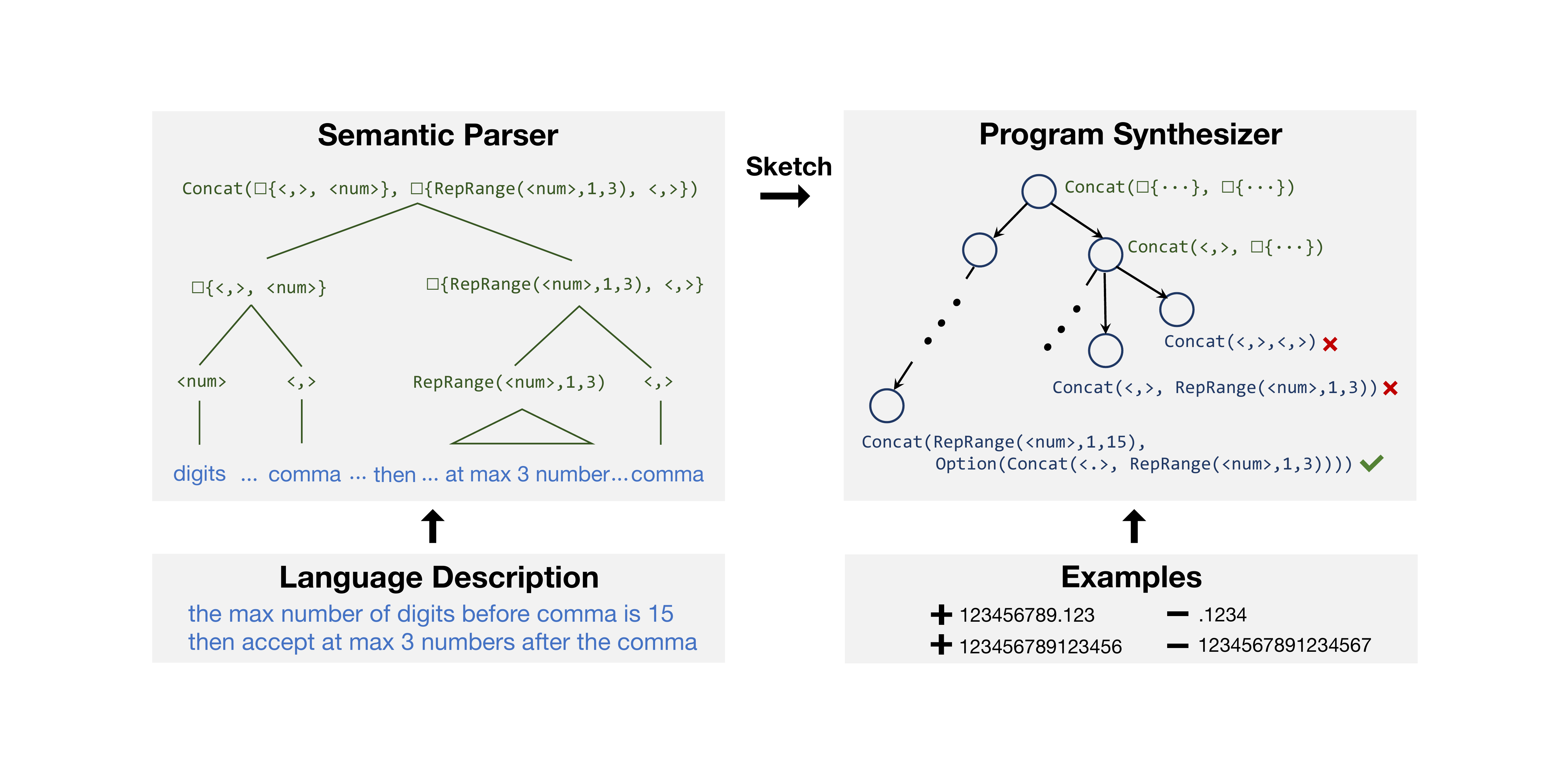}
\centering

\caption{Our regex synthesis approach from language and positive/negative examples. Natural language is parsed into a sketch using a semantic parser. The finished sketch (the root node of the tree) is passed to a program synthesizer, which searches over programs consistent with the sketch and examples. Each leaf node in the search tree is a concrete regex; we return the first one consistent with all the examples.}
\figlabel{framework}
\vspace{-0.1in} 
\end{figure*}

Regular expressions (regexes) are widely used in various domains, but are notoriously difficult to write: \texttt{regex} is one of the most popular tags of posts on Stack Overflow,
with over 200,000 posts. Recent research has attempted to build semantic parsers that can translate natural language descriptions into regexes, via rule-based techniques \cite{Ranta:98}, semantic parsing \cite{KB13}, or seq-to-seq neural network models \cite{deepregex,semregex,softregex}. However, while this prior work has achieved relatively high accuracy on benchmark datasets, trained models still do not generalize to real-world applications: these benchmarks describe simple regexes with short natural language descriptions and limited vocabulary.


Real-world regexes are more complex in terms of length and tree-depth, requiring natural language descriptions that are longer and more complicated \cite{zexuan18}. Moreover, these descriptions may be under-specified or ambiguous. One way to supplement such descriptions is by including positive/negative examples of strings for the target regex to match. In fact, such examples are typically provided by users posting questions on Stack Overflow. Previous methods cannot leverage the guidance of examples at test time beyond naive postfiltering.

In this paper, we present a framework to exploit both natural language and examples for regex synthesis by means of a \emph{sketch}. Rather than directly mapping the natural language into a concrete regex, we first parse the description into an intermediate representation, called a sketch, which is an incomplete regular expression that contains holes to denote missing components. This representation allows our parser to recognize partial structure and fragments from the natural language without fully committing to the regex's syntax. We then use an off-the-shelf program synthesizer, mildly customized for our task, to produce a regex consistent with both the sketch and the provided examples. Critically, this two-stage approach modularizes the language interpretation and program synthesis, allowing us to freely swap out these components.

We evaluate our framework on several English datasets. Because these datasets vary in scale, we consider two sketch generation approaches: neural network-based with a seq-to-seq model \cite{attention} or grammar-based with a semantic parser \cite{sempre}. We use two large-scale datasets from past work, the \kb{} dataset of \newcite{KB13} and the \turk{} dataset of \newcite{deepregex}, augmented with automatically-produced positive/negative examples for each regex. 
Our neural sketch model can exploit these large labeled datasets, allowing our sketch-driven approach to outperform existing seq-to-seq methods, even when those methods are modified to take advantage of examples.
                             
To test our model in a more realistic setting, we also evaluate on a dataset of real-world regex synthesis problems from Stack Overflow. These problems organically have English language descriptions and paired examples that the user wrote to communicate their intent. This dataset is small, only 62 examples; to more robustly handle this setting without large-scale training data, we instantiate our sketch framework with a grammar-based semantic parser. Our approach can solve 57\% of the benchmarks, where existing deep learning approaches solve less than 10\%. While more data is needed, this dataset can motivate further work on more challenging regex synthesis problems.

\section{Regex Synthesis Framework}

In this section, we illustrate how our regex synthesis framework works using a real-world example from a Stack Overflow post.\footnote{https://stackoverflow.com/questions/19076566/regular-expression-that-validates-decimal-18-3} In this post, the user describes the desired regex as ``\emph{the max number of digits before comma is 15 then accept at max 3 numbers after the comma.}'' Additionally, the user provides eight positive/negative examples to further specify their intent. In this instance, the NL description is under-specified: the description doesn't clearly say whether the decimal part is compulsory, and a period (.) is mistakenly described as a comma. These issues in NL pose problems for systems attempting to directly generate the target regex based only on the description.

\figref{framework} shows how our framework handles this example. The natural language description is first parsed into a sketch by a semantic parser, which in this case is grammar-based (Section~\ref{sec:grammar-parser}) but could also be neural in nature (Section~\ref{sec:neural-parser}).
The purpose of the sketch is to capture useful components from the description as well as the high-level structure of the regex. For example, the sketch in \figref{framework} depicts the target regex as the concatenation of two regexes, where the first regex likely involves composition of \regdsl{<num>} and \regdsl{<,>} in some way. We later feed this sketch, together with positive/negative examples, into the synthesizer, which enumeratively instantiates holes with constructs from our regex DSL until a consistent regex is found.

We describe our semantic parsers in Section~\ref{sec:parser} and our synthesizer in Section~\ref{sec:synthesizer}.

\paragraph{Regex/Sketch DSL}

\begin{figure}
    \centering
    \small
    \begin{align*}
        \textcolor{black}{S :=}&\textcolor{black}{  \ C  \ | \  \regdsl{StartsWith}(S) \ | \ \regdsl{EndsWith}(S) }\\
        & \textcolor{black}{ \ | \ \regdsl{Contains}(S) \ | \ \regdsl{Optional}(S)} \\
        & \textcolor{black}{ \ | \ \regdsl{Repeat}(S, k) \ | \ \regdsl{KleeneStar}(S)} \\
        & \textcolor{black}{\ | \ \regdsl{RepAtLeast}(S, k) }\\
        &\textcolor{black}{ \ | \ \regdsl{RepRange}(S, k_1, k_2) }\\
        &\textcolor{black}{ \ | \ \regdsl{Concat}(S, S) \ | \ \regdsl{And}(S, S) \ | \ \regdsl{Or}(S, S) } \\
        & \textcolor{red}{\ | \ \hole\{S,\ldots,S\}}
    \end{align*}
    \caption{Regex DSL (black) and Sketch DSL (all rules including the last rule in red). $C$ represents either a character class such as \regdsl{<let>}, \regdsl{<num>} or a single character such as \regdsl{<a>}, \regdsl{<1>}. $k$ represents an integer. }
    \label{fig:regex_dsl}
\end{figure}

Our regex language (Figure~\ref{fig:regex_dsl}) is similar to the one presented in \cite{deepregex} but more expressive. Our DSL adds some additional constructs, such as \regdsl{Repeat(S,k)}, repeating a regex \regdsl{S} exactly \regdsl{k} times, in our DSL, which is not supported by \citet{deepregex}. This DSL is equivalent in power to standard regular expressions, in that it can match any regular language. 

Our sketch language builds on top of our regex DSL by adding a new construct called a ``constrained hole'' (the red rule in Figure~\ref{fig:regex_dsl}).
Our sketch DSL introduces an additional grammar symbol $\sketch$ and the notion of hole $\hole$.
Holes can be produced with the rule $\hole\{S_1,\ldots,S_m\}$, where each $S_i$ on the right hand side is also a sketch.
A concrete regex $\prog$ belongs to the space of regexes defined by a constrained hole if at least one of its $S_i$ defines any concrete regex $r'$ that is one of the subtrees in $r$. 
Put another way, the regex rooted at $S$ must contain a subtree that matches at least one of the $S_i$, but it does not have to match all of them. However, the synthesizer we use supports using the $S_i$ in its search heuristic to prefer certain programs. In this fashion, the constraint serves as a hint for the leaf nodes of the regex, but it only loosely constraints the structure.

For example, consider the sketch shown in \figref{framework}. Here,  all programs on the leaf nodes of the search tree are included in the space of regexes defined by this sketch.
Note that the first two explored regexes only include some of the components mentioned in the sketch (e.g., $\texttt{<num>}$ and $\texttt{RepRange(<num>,1,3)}$), whereas the final correct regex happens to include every mentioned component. 

\paragraph{Use of holes} There is no single correct sketch for a given example. A trivial sketch consisting of just a single hole could synthesize to a correct program if the examples precisely specify the semantics; this reduces to a pure programming-by-example setting. A sketch could also make no use of holes and fully specify a regex, in which case the synthesizer has no flexibility in its search.

Our process maintains uncertainty over sketches in both training, by leaving them as latent variables, and test, by feeding a $k$-best list of sketches into our synthesizer. In practice, we observe a few patterns in how sketches are used. One successful pattern is when sketches balance concreteness with flexibility, as shown in \figref{framework}: they commit to some high-level details to constrain the search space while using holes with specified components further down in the tree. A second pattern we observe is when the sketch has a single hole at the root but enumerates a rich set of components $S_i$ that are likely to appear; this prefers synthesizing sketches using these subtrees.

\section{Semantic Parser}
\label{sec:parser}

Given a natural language description $\lang=l_1,l_2,\ldots,l_m$, our semantic parser generates a sketch $\sketch$ that encapsulates the user's intent. When combined with examples in the synthesizer, this sketch should yield a regex matching the ground truth regex. As stated before, our semantic parser is a modular component of our system, so we can use different parsers in different settings. We investigate two paradigms of semantic parser: a seq-to-seq neural network parser and a grammar-based parser, as well as two ways of training the parser: maximum likelihood estimation based on a pseudo-gold sketch and maximum marginal likelihood based on whether the sketch leads to the correct synthesis result.

\subsection{Neural Parser}
\label{sec:neural-parser}


\begin{figure*}[t]
\centering
\includegraphics[width=\textwidth, trim=20 105 20 100,clip]{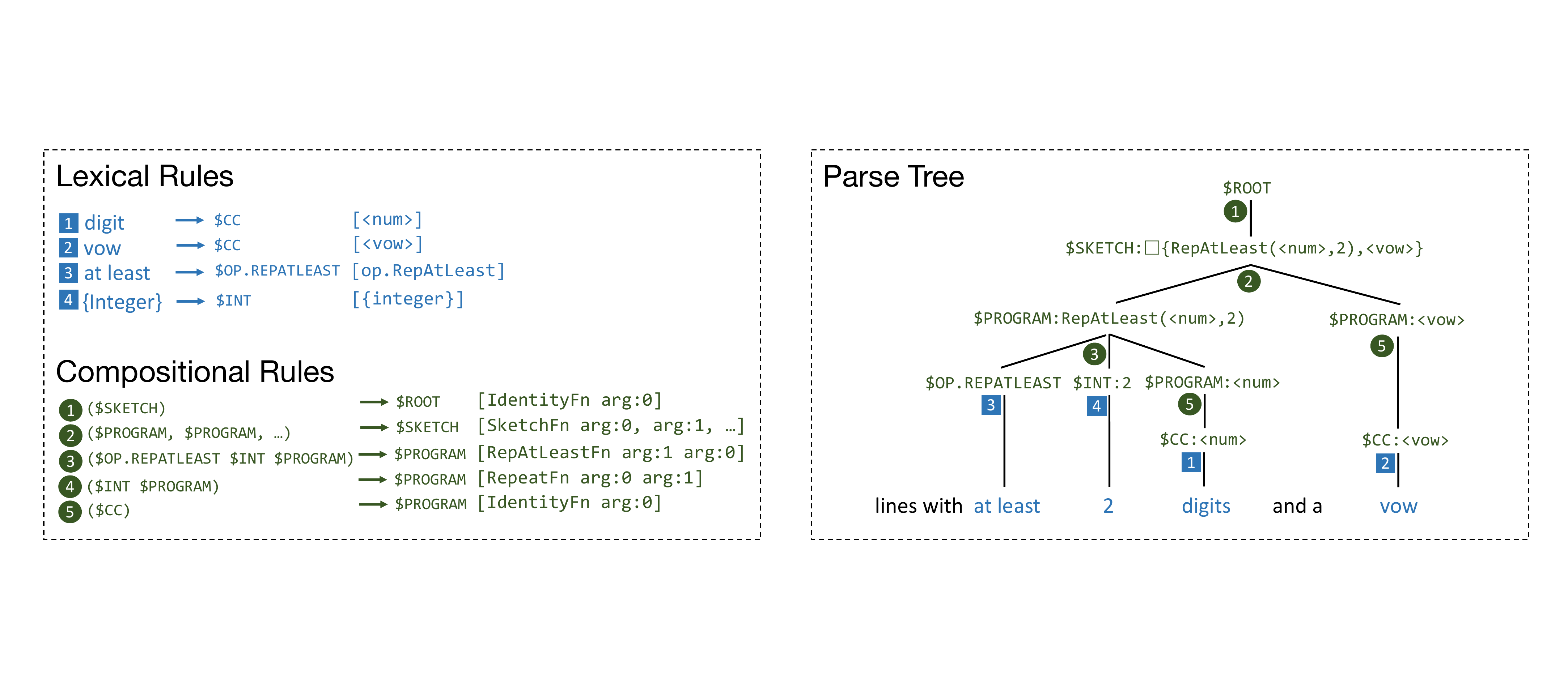}
\centering
\caption{ Examples of rules and the parse tree for building one possible derivation. The left side of a rule is the source sequence of tokens or syntactic categories (marked with a \$ sign). The right side specifies the target syntactic category and then the target derivation or a semantic function (together with arguments) producing it. \textsc{\$program} denotes a concrete regex without holes and \textsc{\$sketch} denotes sketches containing holes. Lexical rule 4 denotes mapping any token of an integer to its value.}
\figlabel{parse_tree}
\end{figure*}

Following recent work \cite{deepregex}, we employ a seq-to-seq model with attention \cite{attention} as our neural parser. Here, we treat the sketch as a sequence of tokens $\sketch=s_1,s_2,\ldots,s_n$ and model $P(\sketch|\lang)$ autoregressively.

Our encoder is a single-layer bidirectional LSTM, where the tokens $l_i$ in the natural language description are encoded into a sequence of hidden states $\bh_i$.
Our decoder is a single-layer unidirectional LSTM, initialized with the encoder final state $\bh_m$. At each timestep $t$, we concatenate the decoder hidden state $\hh_{t}$ with a context vector $c_t$ computed based on bilinear attention, and the probability distribution of the output token $s_t$ is given as:
\begin{align*}
    a_{i,t} = \mathrm{softmax}(\bh_i^\top W_q \hh_t) \ \ \ \ \ \ \ c_t = \sum_i a_{i,t} \bh_i \\
    p(s_t|\lang,s_{<t}) =\mathrm{softmax}(W_z[\hh_t;c_t]),
\end{align*}
where $\hw_i$ is the embedded word vector of $z_i$. The final probability for a generating $S$ conditioned on $L$ is given as $p(S|L)= \prod_{t=1}^{n} p(s_t|\lang,s_{<t})$. 
\subsection{Grammar-Based Parser}
\label{sec:grammar-parser}

We also explore a grammar-based semantic parser built using \sempre{} \cite{sempre}.
This approach is less data hungry than deep neural networks and promises better generalizability, as it is regulated by a grammar and has fewer parameters, which makes it less likely to fit annotation artifacts of crowdsourced datasets.

Given a natural language description, our semantic parser uses a grammar to construct possible sketches. The grammar consists of two sets of rules, lexical rules and compositional rules. Formally, a grammar rule is of the following form:
$\alpha_1 ... \alpha_n \rightarrow  c[\beta]  $. Such a rule maps the sequence of tokens or syntactic categories $\alpha_1 ... \alpha_n$ into target derivation $\beta$ with syntactic category $c$.

As shown in \figref{parse_tree}, each lexical rule maps a word or a phrase in the description to a base concept in the DSL, including character classes, string constants, and operators. Compositional rules generally capture the higher-level DSL constructs, specifying how to combine one or more base concepts to build more complex ones. Our semantic parser constructs possible derivations of sketches by recursively applying these rules, first generating derivations for spans matching the lexical rules and then combining these with compositional rules. Finally, we take the the derivations over the entire natural language description with a designated \textsc{\$root} category as the final set of output sketches.

We design our grammar\footnote{A readable version of grammar is available at \url{https://github.com/xiye17/SketchRegex/blob/master/readable_grammar.pdf}.} according to our sketch DSL. For all the datasets in evaluation, we use a unified grammar that consists of approximately 70 lexical rules and 60 compositional rules. The size of grammar is reflective of the size of DSL, since either a terminal of a single DSL construct needs several rules to specify it (e.g., both \textit{digit} or \textit{number} can present \texttt{<num>}, and \texttt{Concat(X,Y)} can be described in multiple ways like \textit{X before Y} or \textit{Y follows X}). Despite the fact that the grammar is hand-crafted, it is sufficient to cover the fairly narrow domain of regex descriptions.

Our parser allows skipping arbitrary tokens (\figref{parse_tree}), resulting in a large number of derivations. We define a log-linear model to place a distribution over derivations $Z\in \mathcal{D}(\lang)$ given description $\lang$: 
$p_{\theta}(Z|\lang) = \frac{\exp (\theta^\top \phi(\lang,Z))}{\sum_{Z'\in \mathcal{D}(\lang)}\exp (\theta^\top \phi(\lang,Z'))}$
where $\theta$ is the vector of parameters to be learned, and $\phi(\lang,Z)$ is a feature vector extracted from the derivation and description. The features used in our semantic parser are standard features in the \sempre{} framework and mainly characterize the relation between description and applied composition, including indicators when rule $r$ is fired over a span containing token $l$, indicators of whether a particular rule $r$ is fired, and indicators of rule bigrams in the tree.

\subsection{Training}
\label{sec:training}
For both the neural and grammar-based parsers, we can train the model parameters in two ways.
\paragraph{MLE}
Maximum likelihood estimation maximizes the probability of mapping the description to a corresponding gold sketch $S^*$:
$$\argmax_{\theta} \sum_{(S^*,L)} \log p_\theta(S^* \mid \lang).$$
Gold sketches are not defined a priori; however, we describe ways to heuristically derive them in Section~\ref{sec:preprocessing}.
\paragraph{MML}
For a given natural language description and regex pair, multiple syntactically different sketches can yield semantically equivalent regexes. We can therefore maximize the marginal likelihood of generating a sketch that leads us to the semantically correct regex, instead of a generating a particular gold sketch. Namely, we learn the parameters by maximizing:
$$\argmax_{\theta}\sum_{(r^*,L)} \log \sum_{S} 1[\textrm{synth}(S)=r^*] p_\theta(S \mid \lang)$$
where $r^*$ is the ground truth regex and $\textrm{synth}$ denotes running the synthesizer. Computing the sum over all sketches is intractable, so we sample sketches from beam search to approximate the gradients \cite{guu17}.

\section{Program Synthesizer} \label{sec:synthesizer}

In this section, we describe the program synthesizer which takes as input a sketch and a set of examples and returns a regex that is consistent with the given examples. 
Specifically, our synthesizer explores the space of programs defined by the sketch while additionally being guided by the examples. 

\paragraph{Enumerative Synthesis from Sketches} We use an enumeration-based program synthesizer that is a generalized version of the regex synthesizer proposed by \newcite{lee}. Given a program sketch and a set of positive/negative examples, the synthesizer searches the space of programs that can be instantiated by the given sketch and returns a concrete regex that accepts all positive examples and rejects all negative examples.
        

Specifically, the synthesizer instantiates each hole with our DSL constructs or the components for the hole. 
If a hole is instantiated with a DSL terminal such as \texttt{<num>} or \texttt{<let>}, the hole will just be replaced by the terminal. 
If a hole is instantiated using a DSL operator, the hole will first be replaced by this operator, we introduce new holes for its arguments, and we require the components for at least one of the holes to be the original holes' components. 
See \figref{framework} for an example of regexes that could be instantiated from the given sketch. 

Whenever the synthesizer produces a complete instantiation of the sketch (i.e., a concrete regex with no holes), it returns this regex if it is also consistent with the examples (accepts all positive and rejects all negative examples). Otherwise, the synthesizer moves on to the next program in the sketch language. The synthesizer terminates when it either finds a regex consistent with the examples or it has exhausted every possible instantiation of the sketch up to depth $d$.

Our synthesizer differs from that of \newcite{lee} in two main ways. 
First, their regex language is extremely restricted, only allowing the characters 0 and 1 (a binary alphabet). Second, their technique enumerates DSL programs from scratch, whereas our synthesizer performs enumeration based on an initial sketch. This significantly reduces the search space and therefore allows us to synthesize complex regexes much more quickly.

\paragraph{Enumeration Order} Our  synthesizer maintains a worklist of partial programs to complete, and enumerates complete programs in increasing order of depth. Specifically, at each step, we pop the next partial program with the highest overlap with our sketch, expand the hole given possible completions, and add the resulting partial programs back to the worklist. When a partial program is completed (i.e., no holes), it is checked against the provided examples.  The program will be returned to the user if it is consistent with all the examples, otherwise the worklist algorithm continues. 

Note that in this search algorithm, constrained holes are not just hard constraints on the search but are also used to score partial programs, favoring programs using more constructs derived from the natural language. This scoring helps the model prioritize programs that are more congruent with the natural language, lead to more accurate synthesis.

\section{Datasets}
\label{sec:datasets}

We evaluate our framework on two datasets from prior work, \kb\ and \turk, and a new dataset, \so. Statistics about these datasets are given in Table~\ref{tbl:datastats}, and we describe them in more detail below. Since our framework requires string examples which are absent in the existing datasets, we introduce a systematic way to generate positive/negative examples from ground truth regexes.

\paragraph{KB13} \kb{} \cite{KB13} was created with crowdsourcing in two steps. First, workers from Amazon Mechanical Turk wrote the original English language descriptions to describe a subset of the lines in a file. Then a set of programmers from oDesk are required to write the corresponding regex for each of these language descriptions. In total, 834 pairs of description and regex are generated.

\begin{table}[t!]
\small
\centering
\begin{tabular}{l|c|c|c}
\toprule
     Dataset & \textsc{KB13} & \textsc{Turk} & \textsc{SO} \\
\midrule
  size & 824 & 10,000 & 62 \\
  \#. unique words & 207 & 557 & 301 \\
  Avg. NL length & 8.1 & 11.5 & 25.4 \\
  Avg. regex size & 5.1 & 4.9 & 13.2 \\
  Avg. regex depth & 2.5 & 2.3 & 4.0 \\
 \bottomrule
\end{tabular}
\caption{ Statistics of our datasets. Compared to \kb{} and \turk{}, \so{} contains more sophisticated descriptions and regexes. }
\label{tbl:datastats}
\end{table}

\paragraph{Turk} \newcite{deepregex} collected the larger-scale \turk{} dataset to investigate the performance of deep neural models on regex generation. Since it is challenging and expensive to hire crowd workers with domain knowledge, the authors employ a generate-and-paraphrase procedure instead. Specifically, 10,000 instances are randomly sampled from a predefined manually crafted grammar that synchronously generates both regexes and synthetic English language descriptions. The synthetic descriptions are then paraphrased by workers from Mechanical Turk.

The generate-and-paraphrase procedure is an efficient way to obtain description-regex pairs, but it also leads to several issues that we find in the dataset. The paraphrase procedure inherently limits the originality in natural language, leading to artificial descriptions. In addition, since the regexes are stochastically generated without being validated, many of them are syntactically correct but semantically meaningless. For instance, the regex \texttt{\textbackslash b(<vow>)\&(<num>)\textbackslash b} for the description \textit{lines with words containing a vowel and a number} is a valid regex but does not match any string values. These null regexes account for around 15\% of the data. Moreover, other regexes have formulaic descriptions since their semantics are randomly made up (more examples can be found in Section \ref{sec:exp}).

\begin{figure}[t]
\small
    \begin{tabularx}{\linewidth}{|X|}
    \hline
        \bf \kb{} \\
        \textit{lines where there are two consecutive capital letters} \\
         \bf \turk{}  \\
         \textit{lines where words include a digit, upper-case letter, plus any letter} \\
         \bf \so{} \\
        \textit{I'm looking for a regular expression that will match text given the following requirements: contains only 10 digits (only numbers); starts with ``9''} \\
    \hline
    \end{tabularx}
    \caption{ Examples of natural language description from each of the three datasets. \turk{} tends to be very formulaic, while \so{} is longer and much more complex.}
    \label{fig:my_label}
\end{figure}

\subsection{StackOverflow}

To explore regex generation in real-word settings, we collect a new dataset consisting of posts on Stack Overflow. We search posts tagged as regex on Stack Overflow and then filter the collected posts with two rules: (1) the post should include both an English language description as well as positive/negative examples; (2) the post should not contain \emph{abstract concepts} (e.g., ``months'', ``US phone numbers'') or \emph{visual formatting} (e.g., ``AB-XX-XX'') in description.
We collected 62 posts\footnote{These posts are filtered from roughly 1000 top posts. Despite the fact that more data is available on the website, we only view the top posts because the process requires significant human involvement.} that contain both description and regex using our rules. In addition, we slightly preprocess the description by fixing typos and marking string constants, as what's done in prior datasets \cite{deepregex}.

Although \so{} only includes 62 examples, the number of unique words in the dataset is higher than that in KB13 (Table~\ref{tbl:datastats}). Moreover, its average description length and regex size are substantially higher than those of previous datasets, which indicates the complexity of regexes used in real-world settings and the sophistication of language used to describe them.

\subsection{Dataset Preprocessing} 
\label{sec:preprocessing}

\paragraph{Generating Positive/Negative Examples} The \so{} dataset organically has positive/negative examples, but, for the other datasets, we need to generate examples to augment the existing datasets. We use the automaton library \cite{automaton} for this purpose. For positive examples, we first convert the ground truth regex into an automaton and generate strings by sampling values consistent with paths leading to accepting states in the automaton. For negative examples, we take the negation of the ground truth regex, convert it into an automaton, and follow the same procedure as generating the positive examples. To ensure a diverse set of examples, we limit the number of times that we visit each transition so that the example generator avoids taking the same transitions repeatedly. For each of these datasets, we generate 10 positive and 10 negative examples. This is comparable to what was used in past work \cite{semregex} and it is generally hard to automatically generate a smaller set of ``corner cases'' that humans would write.

\paragraph{Generating Heuristic Sketches}
Our approach does not require any notion of a gold sketch and can operate from weak supervision only. However, we can nevertheless derive \emph{pseudogold} sketches using a heuristic and train with MLE to produce these in order to examine the effects of injecting human prior knowledge into learning. We generate pseudogold sketches from ground truth regexes as follows. 
For any regex whose Abstract Syntax Tree (AST) has depth more than 1, we replace the operator at the root with a constrained hole and the components for this hole are arguments of the original operator.
For example, the gold sketch for regex $\texttt{concat}(\texttt{<num>},\texttt{<let>})$ is $\hole\{\texttt{<num>},\texttt{<let>}\}$. For regexes with depth 1, we just wrap the ground truth regex within a constrained hole; e.g., the gold sketch for the regex $\texttt{<num>}$ is $\hole\{\texttt{<num>}\}$. We apply this method to \turk{}  and \kb{}.

For the smaller \so{} dataset, we explored a more heavily supervised approach where we manually labeled gold sketches based on information from the gold sketch that we judged to be unambiguous about the ground truth regex.
For example, the description ``\emph{The input box should accept only if either (1) first 2 letters alpha + 6 numeric or (2) 8 numeric}'' is labeled with the sketch
 \regdsl{Or($\hole$\{Repeat(<let>,2),Repeat(<num>,6)\}, $\hole$\{Repeat(<num>),8))\}},
 which clearly reflects both the user's intent and the compositional structure.

\begin{table*}[t]
\small
\centering
\begin{tabular}{l|cc|cc}
\toprule
    \multirow{2}{*}{} &\multicolumn{2}{c|}{\kb{}} &\multicolumn{2}{c}{\turk{}} \\
    & Acc & Consistent & Acc & Consistent\\
  \midrule
  \bf  Prior Work: & & & & \\
  \qquad \textsc{DeepRegex} (Locascio et al.)  & 65.6\% & $-$ & 58.2\% & $-$\\
  \qquad \textsc{SemRegex}   & 78.2\% & $-$ & 62.3\% & $-$ \\
  \midrule
  \bf Translation-Based Approaches: & & & & \\ 
 \qquad \textsc{DeepRegex\textsuperscript{MLE}} & 66.5\% & $-$ & 60.3\% & $-$\\
  \qquad \textsc{DeepRegex MLE} & 66.5\% & $-$ & 60.3\% & $-$\\
 \qquad \textsc{DeepRegex\textsuperscript{MML}} & 68.2\% & $-$ & 62.4\% & $-$\\
  \qquad \textsc{DeepRegex\textsuperscript{MLE} + Filter} & 77.7\% & 89.0\% & 82.8\% & 92.0\% \\
\qquad \textsc{DeepRegex\textsuperscript{MML} + Filter} & 80.1\% & 91.7\%  & 84.3\% & 92.8\% \\

 \midrule
 \midrule
  \bf Sketch-Driven (No Training): &  & &  & \\
  \qquad \textsc{Empty Sketch}  &  15.5\% & 18.4\% & 21.0\% & 34.4\%\\
  \qquad \gramsketch{} \textsc{(Max Coverage)} & 68.0\% & 76.7\% & 60.2\% & 78.8\%\\ 
  \midrule
  \bf Sketch-Driven (No Sketch Supervision): & & & & \\
  \qquad \textsc{DeepSketch\textsuperscript{MLE}}& 76.2\% & 88.8\% & 74.6\% & 92.8\% \\
  \qquad \textsc{DeepSketch\textsuperscript{MML}}  & 82.5\% & 94.2\% & 84.3\% & 95.8\% \\   
  \midrule
 \bf Sketch-Driven (Pseudogold Sketches): &  & &  & \\
  \qquad \gramsketch{} & 72.8\% & 85.4\% & 69.4\% & 87.4\%\\ 
  \qquad \textsc{DeepSketch\textsuperscript{MLE} pseudogold}  &  84.0\% & 95.3\% &  85.4\% & 98.4\% \\
  \qquad \textsc{DeepSketch\textsuperscript{MML} pseudogold}  & \bf 86.4\% & 96.3\% & \bf 86.2\% & 98.9\%\\    
 
  \bottomrule
\end{tabular}
\caption{Results on datasets from prior work. We evaluate on both accuracy (Acc) and the fraction of regexes produced consistent (Consistent) with the positive/negative examples. Our sketch-driven approaches outperform prior approaches even when those are modified to use examples. Our approach can leverage heuristic pseudogold sketches, but does not require them. Our \textsc{DeepSketch} models achieve the best results, but even our grammar-based method (\gramsketch{}) outperforms past systems that do not use examples.}
\label{tbl:public-dataset}
\end{table*}

\section{Experiments}
\label{sec:exp}

\paragraph{Setup} We implement all neural models in \textsc{Pytorch} \cite{pytorch}. While training with MLE, we use the Adam optimizer \cite{adam} with a learning rate of 1e-3 and a batch size of 25. We train our models until the loss on the development set converges. When training with \textsc{MML}, we set the learning rate to be 1e-4 and use beam search with beam size 10 to approximate the gradients.

We build our grammar-based parsers on top of the SEMPRE framework \cite{sempre}. We use the same grammar for all three datasets. On datasets from prior work, we train our learning-based models with the training set. On \so{}, we use 5-fold cross-validation as described in Section \ref{sec:datasets} because of the limited data size. Our grammar-based parser is always trained for 5 epochs with a batch size of 50 and use a beam size of 200 when trained with MML.

During the testing phase, we produce a $k$-best list of sketches for a given NL description and run the synthesizer on each sketch in parallel. For a single sketch, the synthesizer either finds an example-consistent regex or running out of a specified time budget (timeout). We pick the output of the highest-ranked sketch yielding an example-consistent regex as the answer.
For \kb{} and \turk{}, we set the beam size $k$ to be 20 and set the timeout of synthesizer to be 2s.
For the more challenging \so{} dataset, we synthesize top 25 sketches and set the timeout to be 30s. In the experiments, the average time to synthesize a single sketch for a single benchmark in \turk{}, \kb{}, and \so{} is 0.4s, 0.8s, and 10.8s, respectively, and we synthesize the $k$-best lists in parallel using 10 threads.


\subsection{Evaluation: \kb{} and \turk{}} 
\label{sec:result_turk}

\paragraph{Baselines: Prior Work + Translation-based Approaches}

We compare our approach against several baselines. \deepregex{} directly translates language descriptions with a seq-to-seq model without looking at the examples using the \textsc{MLE} objective. Note that we compare against both reported numbers from \citet{deepregex} as well as our own implementation of this (\deepregex{}\textsuperscript{MLE}), which outperforms the original by 0.9\% and 2.0\% on \kb{} and \turk{}, respectively; we use this version in all other reported experiments.

\semregex{} \cite{semregex}\footnote{Upon consultation with the authors of \semregex{} \cite{semregex}, we were not able to reproduce the results of their model. Therefore, we only include the printed numbers of semantic accuracy on the prior datasets.} uses the same model as \deepregex{} but is trained to maximize semantic correctness of the gold regex, rather than having to produce an exact match. We implement a similar technique using maximum marginal likelihood training to optimize for semantic correctness (\deepregex{}\textsuperscript{MML}).

Note that none of these methods assumes access to examples to check correctness at \emph{test} time. To compare these methods to our setting, we extend them in order to exploit examples: we produce the model's $k$-best list of solutions, then take the highest element in the $k$-best list consistent with the examples as the answer. We apply this method to both types of training to yield \textsc{DeepRegex\textsuperscript{MLE}+Filter} and \textsc{DeepRegex\textsuperscript{MML}+Filter}.

\paragraph{Sketch-Driven} We evaluate three broad types of our sketch-driven models.

Our \textbf{No Training} approaches only use untrained sketch procedures. As an example-only baseline, we include the results using an \textsc{Empty Sketch} (a single hole), relying entirely on the synthesizer. We also use a variant of \textsc{GrammarSketch} method where we heuristically prefer sketch derivations that cover as many words in the input sentence as possible (\textsc{GrammarSketch (Max Coverage)}).

In the \textbf{No Sketch Supervision} setting, we assume no access to labeled sketch data. However, it is challenging to train a neural sketch parser from randomly initialized parameters purely with the \textsc{MML} objective. We therefore warm start the neural models using \textsc{GrammarSketch (Max Coverage)}: we rank the sketches by their coverage of the input sentence, and take the highest-coverage sketch which synthesizes to the correct ground truth regex (if one can be found) as a gold sketch for warm-starting. We can train with the MLE objective for a few epochs and then continue with MML training (\textsc{DeepRegex\textsuperscript{MML}}). As a comparison, we can also evaluate the model trained only with MLE with these sketches (\textsc{DeepRegex\textsuperscript{MLE}}).

Models in the \textbf{Pseudogold Sketches} setting follow the approach described in the previous paragraph, but uses the pseudogold sketches described in Section~\ref{sec:preprocessing} instead of bootstrapping with the grammar-based approach.

\paragraph{Results}
Table \ref{tbl:public-dataset} summarizes our experimental results on these two datasets. Note that reported accuracy is semantic accuracy, which measures the functional equivalence of the regex compared to the ground truth. First, we find a significant performance boost by filtering the output of our \deepregex{} variants using examples (11.2\% on \kb{} and 21.5\% on \turk{} when applying this to \textsc{DeepRegex\textsuperscript{MLE}}), indicating the utility of examples in verifying the produced regexes.

However, our sketch-driven approach outperform these previous approaches even when they are extended to benefit from examples. We achieve new state-of-the-art results on both datasets, with slightly stronger performance when pseudogold sketches are used. 
The results are particularly striking in terms of consistency (fraction of regexes produced consistent with the examples). Because we allow uncertainty in the sketches and use examples to guide the construction of regexes, our framework achieves 50\% or more relative reduction in the rate of inconsistent regexes compared to \drex{} baseline (91.7\% and 92.8\% on the two datasets), which may fail if no consistent sketch is in the $k$-best list.

We also find that our \gramsketch{} approach, trained with pseudogold sketches, achieves nearly 70\% accuracy on both datasets, which is better than \deepregex{}. This indicates the generalizability of this approach. The performance of \gramsketch{} lags that of \drex{} and \deepsketch{} models, which can be attributed to the fact that \gramsketch{} is more constrained by its grammar and is less capable of exploiting large amounts of data compared to neural approaches.


Finally, we turn to the source of the supervision. The untrained \textsc{GrammarSketch(Max Coverage)} achieves over 60\% accuracy on both datasets; recall that this provides the set of gold sketches as initial supervision in our warm-started model.
Our sketch-driven approach trained with \textsc{MML} (\textsc{DeepSketch\textsuperscript{MML}}) achieves 82.5\% on \kb{} and 84.3\% on \turk{}, which is comparable with the performance obtained using pseudogold sketches, demonstrating that human labeling or curation of sketches is not required for this technique to work well.

\subsection{Evaluation: Stack Overflow}

\paragraph{Additional Baselines}
It is impractical to train a deep neural model from scratch on this dataset, so we modify our approach slightly to compare against such models. First, we train a model on \turk{} and fine-tune it on \so{} (Transferred Model). Second, we explore a modified version of the dataset where we rewrite the descriptions in \so{} to make them conform to the style of \turk{} (Curated Language), as users might do if they were knowledgeable about the capabilities of the regex synthesis system they are using. For example, we manually paraphrase the original description \textit{``write regular expression in C\# to validate that the input does not contain double spaces''} to \textit{``line that does not contain `space' two or more times''}, and apply \drex{} method on the curated descriptions (without fine-tuning on them). Note that this simplifies the inputs for these baselines considerably by removing variation in the language.


We also construct a grammar-based regex parser from our \gramsketch{} model by removing the grammar rules related to assembling sketches from regexes. We use our filtering technique as well and call this the \textsc{GrammarRegex+Filter} baseline.



\begin{table}[t!]
\renewcommand{\tabcolsep}{0.88mm}
\small
\centering
  \begin{tabular}{l c c c}
    \toprule
    \multirow{2}{*}{Approach} &
      \multicolumn{3}{c}{Top-N Acc}\\
      & {top-1} & {top-5} & {top-25} \\
      \midrule
    \textsc{\b DeepRegex+Filter}\\
    \; Transferred Model & 0\% & 0\%  & 0\% \\
    \; +Curated Language& 0\% & 0\%  & \phantom{0}6.6\% \\
    \midrule
        \textsc{GrammarRegex+Filter} & \phantom{0}3.2\% & \phantom{0}9.7\%& 11.3\%\\
    \midrule
            \textsc{Empty Sketch} & \phantom{0}4.8\% & $-$ & $-$ \\
    \midrule
    \deepsketch{}\\
    \; Transferred Model & \phantom{0}3.2\% & \phantom{0}3.2\%  & \phantom{0}4.8\%  \\
    \midrule
    \gramsketch{}\\ 
    \; \textsc{Max Coverage}  &  16.1\% & 34.4\% & 45.2\%\\
    \; \textsc{MLE, manual sketches} &  \bf 34.4\% & 48.4\% & 53.2\% \\
    \; \textsc{MML, no sketch sup} & 31.1\% &\bf 54.1\% & \bf 56.5\% \\
    \bottomrule
  \end{tabular}
  \caption{ Results on the \so{} dataset. The \deepregex{} method totally fails even when the examples are generously rewritten to conform to the model's ``expected'' style. Our \gramsketch{} model can do significantly better, with or without manually-labeled sketches.}
  \label{tbl:so}
\end{table}

\paragraph{Results} Since \so{} is a challenging dataset, we report the top-N accuracy, where the model is considered correct if any of the top-N sketches synthesizes to a correct answer. Table~\ref{tbl:so} shows the results on this dataset.
The transferred \textsc{DeepRegex} model completely fails on these real-world tasks. Rewriting and curating the language, we are able to get some examples correct among the top 25 derivations, but only on very simple cases. \textsc{GrammarRegex+Filter} is similarly unable to do well: this approach is too inflexible given the complexity of regexes in this dataset.
Our transferred \deepsketch{} approach isalso  still limited here, as the text is too dissimilar from \turk{}.

Our \gramsketch{} approach, trained without explicit sketch supervision, achieves a top-1 accuracy of 31.1\%. Surprisingly, this is comparable to the performance of \gramsketch{} trained using manually-written sketches, and even outperforms this model in terms of top-5 and top-25 accuracy. This substantially outperforms all of the baseline approaches. We attribute this success to the problem decomposition: because the sketches produced can be simpler than the full regex, our model is much more robust to the complex setting of this dataset. By examining the problems that are solved, we find our approach is able to solve several complicated cases with long descriptions and sophisticated regexes (e.g., the example in \figref{framework}).



%
\subsection{Detailed Analysis}

\paragraph{Data Efficiency}

\begin{figure}[t]
\includegraphics[width=\linewidth,trim=50 242 75 260,clip]{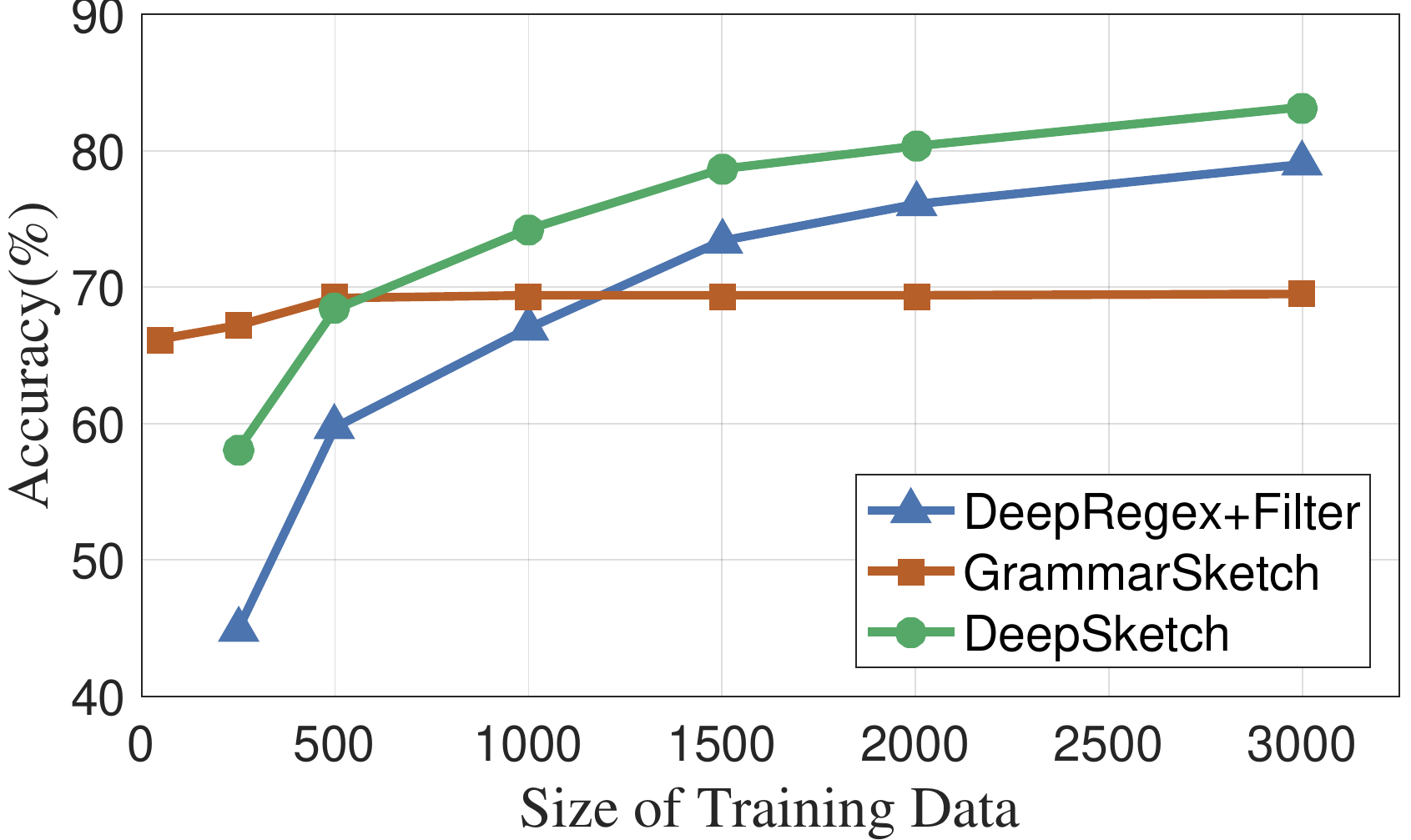}
\centering
\caption{ Accuracy on \turk{} for different training set sizes. Our \deepsketch{} and \gramsketch{} approaches outperform the \drex{} baseline when training data is limited.}
\figlabel{efficiency}
\vspace{-2mm}
\end{figure}

In \figref{efficiency}, we explicitly evaluate the data efficiency of our techniques, comparing the performance of \drex{}, \gramsketch{}, and \deepsketch{} on \turk{}. 
When the data size is extremely limited (no more than 500), our \gramsketch{} approach is much better than the \drex{} baseline. Our \deepsketch{} approach is more flexible and achieves stronger performance for larger training data sizes, consistently outperforming the \drex{} baseline as the training data size increases. More importantly, the difference is particularly obvious when the size of training data is relatively small, in which case correct regexes are less likely to exist in the \deepregex{} generated $k$-best lists due to lack of supervision. By contrast, it is easier to learn the mapping from the language to effective sketches, which are simpler than gold regexes.

Overall, these results indicate that \deepregex{}, as a technique, is only effective when large training sets are available, even for a relatively simple set of natural language expressions.

\paragraph{Impact of Number of Examples}
\begin{table}[t]
    \centering
    \footnotesize
    \begin{tabular}{l c c c c }
    \toprule
    & 4 & 6 & 8 & 10 \\
    \midrule
        \drex{} & 79.5 & 81.0 &	82.3 &	82.8 \\
        \gramsketch{} & 59.4 & 64.4 & 67.6 & 69.5 \\
        \deepsketch{} & 71.8 & 78.6 & 82.7 & 85.4\\
    \bottomrule
    \end{tabular}
    \caption{Performance on \turk{} varying the number of positive/negative examples. Because our synthesizer depends on sufficient examples to constrain the semantics, our sketch-based approaches require a certain number of examples to work well.}
    \label{tbl:impact_exs}
\end{table}
We show how the number of positive/negative examples impacts the performance on \turk{} in Table~\ref{tbl:impact_exs}. Our sketch-driven techniques rely on examples to search for the desired regexes in the synthesizer step, and therefore are inferior to \drex{} when only limited number of examples are provided. However, as more examples are included, our sketch-driven approaches are more effective in taking advantage of the multi-modality than simply filtering the outputs (\drex{}). Note that in the \so{} setting, we evaluate on sets of examples provided by actual users, and find that users typically provide enough examples for our model to be in an effective regime.
 
\paragraph{Impact of Beam Size}


\begin{table}[t]
\renewcommand{\tabcolsep}{1.7mm}
    \centering
    \footnotesize
    \begin{tabular}{l c c c c c}
    \toprule
                & 1 & 3 & 5 & 10 & 20 \\
    \midrule
        \drex{}       & 60.3 & 73.2 & 76.8 & 80.3 & 82.8 \\
        \gramsketch{} & 58.3 & 65.0 & 67.1 & 68.9 & 69.5 \\
        \deepsketch{} & 69.4 & 79.7 & 82.3 & 84.3 & 85.4\\
    \bottomrule
    \end{tabular}
    \caption{Performance on \turk{} under different beam sizes.}
    \label{tbl:impact_beam}
    \vspace{-3mm}
\end{table}

We study the effect of varying beam size on the performance on \turk{} in Table~\ref{tbl:impact_beam}. \deepsketch{} outperforms \drex{} by a substantial gap with smaller beam sizes, as we naturally allow uncertainty using the holes in sketches. Note that using larger beam sizes is important for \deepsketch{} because we feed the entire $k$-best list to the synthesizer instead of the top sketch only.


\begin{figure}[t]
    \scriptsize
    \begin{tabularx}{\linewidth}{| r X|}
    \hline

    \multicolumn{2}{|c|}{\bf  \turk{}} \\
    \hline
         \multicolumn{2}{|l|}{\bf Success:} \\
        (a)\; nl: & \textit{lines with 3 more capital letters} \\
         gt: &  \texttt{(<cap>)\{3,\})(.*)} \\
        (b)\; nl: & \textit{none of the lines should have a vowel , a capital letter , or the string ``dog''} \\
         gt: & \texttt{\textasciitilde((<vow>)|(dog)|(<cap>))} \\
         \multicolumn{2}{|l|}{\bf Failure:} \\
         (c)\; nl: &\textit{lines with ``dog'' or without ``truck'' , at least 7 times} \\
         gt: & \texttt{((dog)|(\textasciitilde(truck)))\{7,\}} \\
         error: &  \texttt{(dog)|(\textasciitilde(truck))} \\
        (d)\; nl: & \textit{lines ending with lower-case letter or not the string ``dog''} \\
         gt: & \texttt{(.*)(([<low>])|(\textasciitilde(dog)))} \\
         error: & \texttt{(([<low>])|(\textasciitilde(dog)))*} \\
    \hline
    \multicolumn{2}{|c|}{\bf \so{}} \\
    \hline
    \multicolumn{2}{|l|}{\bf Success:} \\
        (e)\; nl: & \textit{valid characters are alphanumeric and ``.''(period). The patterns are ``\%d4\%'' and ``\%t7\%''. So ``\%'' is not valid by itself, but has to be part of these specific patterns.} \\
         gt: &\texttt{((<let>|<num>|(.)|(\%d4\%)|(\%t7\%))\{1,\}}\\
        (f)\; nl: & \textit{The input box should accept only if either (1) first 2 letters alpha + 6 numeric or (2) 8 numeric} \\
         gt: &\texttt{(<let>\{2\}<num>\{6\})|(<num>\{8\})} \\

    \multicolumn{2}{|l|}{\bf Failure:} \\
         (g)\; nl: & \textit{I'm trying to devise a regular expression which will accept decimal number up to 4 digits} \\
         gt: &\texttt{(<num>\{1,\})(.)(<num>\{1,4\})} \\
         error: &\texttt{(<num>\{1,\})((.)(<num>\{1,4\}))?} \\
        (h)\; nl: & \textit{the first letter of each string is in upper case} \\
         gt: &\texttt{<cap>(<let>)*(( )<cap>(<let>)*)*} \\
         error: &\texttt{((( )<cap>)|(<let>))\{1,\}} \\

    \hline
    \end{tabularx}
    \caption{ Examples of success and failure pairs from \turk{} and \so{}. On pairs (a) and (b), our \deepsketch{} is robust to the issues existing in natural language descriptions. On pairs (c) and (d), our approach fails due to the unrealistic semantics of the desired regexes. \gramsketch{} succeeds in solving some complex pairs in \so{}, including (e) and (f). However, (g) and (h) fail because of insufficient examples or overly concise descriptions. } 
    \label{fig:analysisexs}
    \vspace{-2mm}
\end{figure}

\subsection{Examples of Success And Failure Pairs}

\paragraph{\turk{}}
We now analyze the output of the  \drex{} and \deepsketch{}. \figref{analysisexs} provides some success pairs that \deepsketch{} solves while \drex{} does not. Examples of success pairs suggest that our approach can deal with under-specified descriptions. For instance, in pair (a) from \figref{analysisexs}, the language is ungrammatical (\emph{3 more} instead of \emph{3 or more}) and also ambiguous: should the target string consist of only capital letters, or could it have capital letters as well as something else? Our approach is able to recover the faithful semantics using sketch and examples, whereas \deepregex{} fails to find the correct regex. In pair (b), the description is fully clear but \deepregex{} still fails because the phrase \emph{none of} rarely appears in the training data. Our approach can solve this pair since it is less sensitive to the description. 

We also give some examples of failure cases for our model. These are particularly common in cases of unnatural semantics. For instance, the regex in pair (c) accepts any string except the string \textit{truck} (because \texttt{\textasciitilde(truck)} matches any string but \textit{truck}). The semantics are hard to pin down with examples, but the correct regex is also artificial and unlikely to appear in real word applications. Our \deepsketch{} fails on this pair since the synthesizer fails to catch the \textit{at least 7 times} constraint when strings that have less than 7 characters can also be accepted (since \emph{without truck} can match the empty string). \deepregex{} is able to produce the ground-truth regex in this case, but this is only because the formulaic description is easy enough to translate directly into a regex.

\paragraph{\so{}} We show some solved and unsolved examples using \gramsketch{} from the \so{} dataset. Our approach can successfully deal with multiple-sentence inputs like pairs (e) and (f).
They both contain multiple sentences with each one describing certain a component or constraint, which seems to be a common pattern of describing real world regexes. Our approach is effective for this structure because the parser can extract fragments from each sentence and hand them to the synthesizer for completion.

Some failure cases are due to lack of corner-case examples. E.g., the description from pair (g) doesn't explicitly specify whether the decimal part is required and there are no corner-case negative examples that provide this clue. Our synthesizer mistakenly treats the decimal part as an option, failing to match the ground truth. In addition, pair (h) is an example in which the natural language description is too concise for the grammar parser to generate a useful sketch.
\section{Related Work}

\paragraph{Other NL and program synthesis}
There has been recent interest in synthesizing programs from natural language. One line of work uses either grammar-based or neural semantic parsing to synthesize programs. Particularly, several techniques have been proposed to translate natural language to SQL queries \cite{sqlizer, iyer2017, suhr2018}, ``if-this-then-that'' recipes \cite{ifttt}, bash commands \cite{nl2bash}, Java expressions \cite{java} and more. Our work is different from prior work in that it utilizes input-output examples in addition to natural language.
While several past approaches use both natural language and examples \cite{spoc, algolisp, zhong2020}, they only use the examples to verify the generated programs, whereas our approach heavily engages examples when searching for the instantiation of sketches to make the synthesizer more efficient. 

Another line of work has focused on exploring which deep learning techniques are most effective for directly predicting programs from natural language. Recent work has built encoder-decoder models to generated logical forms or programs represented by sequences \cite{Lidong16}, and ASTs \cite{Rabin17, yin17, idiom2019, idiom2020}.
However, some of the most challenging code settings such as the Hearthstone dataset \cite{ling2016} only evaluates the produced strings by exact match accuracy or BLEU score, rather than executing the programs on real data as we do.
 
There is also recent work using neural models to generate logical forms employing a coarse-to-fine approach \cite{zettlemoyer2009, kwiatkowski2013, artzi2015, coarse-to-fine, wang-2019-learning}, which first generates an abstract logical form and then concretizes it using neural modules, whereas we complete the sketch via a synthesizer.

\paragraph{Program synthesis from examples} Recent work has studied program synthesis from examples in other domains \cite{flashfill,sygus,fidex,neo}. Similar to prior work \cite{deepcoder, kalyan2018neural,odena2020learning}, we implement an enumeration-based synthesizer to search for the target program, but they use probability distribution of functions or production rules predicted by neural networks to guide the search, while our work relies on sketches.

Our method is closely related to sketch-based approaches \cite{sketch, sketchadapt} in that our synthesizer starts with a sketch. However, we produce sketches automatically from the natural language description whereas traditional sketch-based synthesis \cite{sketch} relies on a user-provided sketch, and our sketches are hierarchical and constrained compared to other neural sketch-based approaches \cite{sketchadapt}. 

\section{Conclusion}
We have proposed a sketch-driven regular expression synthesis framework that utilizes  both natural language  and  examples, and we have instantiated this framework with  both a neural and a grammar-based parser. Experimental results reveal the artificialness of existing public datasets and demonstrate the advantages of our approach over existing research, especially in real world settings. 

\section*{Acknowledgments}

This work was partially supported by NSF Grant IIS-1814522, NSF Grant SHF-1762299, gifts from Arm and Salesforce, and an equipment grant from NVIDIA. The authors acknowledge the Texas Advanced Computing Center (TACC) at The University of Texas at Austin for providing HPC resources used to conduct this research. Thanks as well to our TACL action editor Luke Zettlemoyer and the anonymous reviewers for their helpful comments.

\bibliographystyle{acl_natbib}
\bibliography{tacl2018}

\begin{thebibliography}{42}
\expandafter\ifx\csname natexlab\endcsname\relax\def\natexlab#1{#1}\fi

\bibitem[{{Alur} et~al.(2013){Alur}, {Bodik}, {Juniwal}, {Martin},
  {Raghothaman}, {Seshia}, {Singh}, {Solar-Lezama}, {Torlak}, and
  {Udupa}}]{sygus}
R.~{Alur}, R.~{Bodik}, G.~{Juniwal}, M.~M.~K. {Martin}, M.~{Raghothaman}, S.~A.
  {Seshia}, R.~{Singh}, A.~{Solar-Lezama}, E.~{Torlak}, and A.~{Udupa}. 2013.
\newblock Syntax-guided synthesis.
\newblock In \emph{2013 Formal Methods in Computer-Aided Design (FMCAD)}.

\bibitem[{Artzi et~al.(2015)Artzi, Lee, and Zettlemoyer}]{artzi2015}
Yoav Artzi, Kenton Lee, and Luke Zettlemoyer. 2015.
\newblock Broad-coverage {CCG} semantic parsing with {AMR}.
\newblock In \emph{Proceedings of the Conference on Empirical Methods in
  Natural Language Processing (EMNLP)}.

\bibitem[{Balog et~al.(2017)Balog, Gaunt, Brockschmidt, Nowozin, and
  Tarlow}]{deepcoder}
M~Balog, AL~Gaunt, M~Brockschmidt, S~Nowozin, and D~Tarlow. 2017.
\newblock Deepcoder: Learning to write programs.
\newblock In \emph{Proceedings of the International Conference on Learning
  Representations (ICLR)}.

\bibitem[{Berant et~al.(2013)Berant, Chou, Frostig, and Liang}]{sempre}
Jonathan Berant, Andrew Chou, Roy Frostig, and Percy Liang. 2013.
\newblock Semantic parsing on {F}reebase from question-answer pairs.
\newblock In \emph{Proceedings of the Conference on Empirical Methods in
  Natural Language Processing (EMNLP)}.

\bibitem[{Dong and Lapata(2016)}]{Lidong16}
Li~Dong and Mirella Lapata. 2016.
\newblock Language to logical form with neural attention.
\newblock In \emph{Proceedings of the Annual Meeting of the Association for
  Computational Linguistics (ACL)}.

\bibitem[{Dong and Lapata(2018)}]{coarse-to-fine}
Li~Dong and Mirella Lapata. 2018.
\newblock Coarse-to-fine decoding for neural semantic parsing.
\newblock In \emph{Proceedings of the Annual Meeting of the Association for
  Computational Linguistics (ACL)}.

\bibitem[{Feng et~al.(2018)Feng, Martins, Bastani, and Dillig}]{neo}
Yu~Feng, Ruben Martins, Osbert Bastani, and Isil Dillig. 2018.
\newblock Program synthesis using conflict-driven learning.
\newblock In \emph{Proceedings of the 39th ACM SIGPLAN Conference on
  Programming Language Design and Implementation (PLDI)}.

\bibitem[{Gulwani(2011)}]{flashfill}
Sumit Gulwani. 2011.
\newblock Automating string processing in spreadsheets using input-output
  examples.
\newblock In \emph{Proceedings of the 38th Annual ACM SIGPLAN-SIGACT Symposium
  on Principles of Programming Languages (POPL)}.

\bibitem[{Guu et~al.(2017)Guu, Pasupat, Liu, and Liang}]{guu17}
Kelvin Guu, Panupong Pasupat, Evan Liu, and Percy Liang. 2017.
\newblock From language to programs: Bridging reinforcement learning and
  maximum marginal likelihood.
\newblock In \emph{Proceedings of the Annual Meeting of the Association for
  Computational Linguistics (ACL)}.

\bibitem[{Gvero and Kuncak(2015)}]{java}
Tihomir Gvero and Viktor Kuncak. 2015.
\newblock {Synthesizing Java Expressions from Free-form Queries}.
\newblock In \emph{Proceedings of the 2015 ACM SIGPLAN International Conference
  on Object-Oriented Programming, Systems, Languages, and Applications
  (OOPSLA)}.

\bibitem[{Iyer et~al.(2019)Iyer, Cheung, and Zettlemoyer}]{idiom2019}
Srinivasan Iyer, Alvin Cheung, and Luke Zettlemoyer. 2019.
\newblock Learning programmatic idioms for scalable semantic parsing.
\newblock In \emph{Proceedings of the Conference on Empirical Methods in
  Natural Language Processing (EMNLP)}.

\bibitem[{Iyer et~al.(2017)Iyer, Konstas, Cheung, Krishnamurthy, and
  Zettlemoyer}]{iyer2017}
Srinivasan Iyer, Ioannis Konstas, Alvin Cheung, Jayant Krishnamurthy, and Luke
  Zettlemoyer. 2017.
\newblock Learning a neural semantic parser from user feedback.
\newblock In \emph{Proceedings of the Annual Meeting of the Association for
  Computational Linguistics (ACL)}.

\bibitem[{Kalyan et~al.(2018)Kalyan, Mohta, Polozov, Batra, Jain, and
  Gulwani}]{kalyan2018neural}
Ashwin Kalyan, Abhishek Mohta, Oleksandr Polozov, Dhruv Batra, Prateek Jain,
  and Sumit Gulwani. 2018.
\newblock Neural-guided deductive search for real-time program synthesis from
  examples.
\newblock In \emph{International Conference on Learning Representations
  (ICLR)}.

\bibitem[{Kingma and Ba(2015)}]{adam}
Diederik~P Kingma and Jimmy Ba. 2015.
\newblock Adam: A method for stochastic optimization.
\newblock In \emph{Proceedings of the International Conference on Learning
  Representations (ICLR)}.

\bibitem[{Kulal et~al.(2019)Kulal, Pasupat, Chandra, Lee, Padon, Aiken, and
  Liang}]{spoc}
Sumith Kulal, Panupong Pasupat, Kartik Chandra, Mina Lee, Oded Padon, Alex
  Aiken, and Percy~S Liang. 2019.
\newblock Spoc: Search-based pseudocode to code.
\newblock In \emph{Proceedings of the Conference on Advances in Neural
  Information Processing Systems (NeurIPS)}.

\bibitem[{Kushman and Barzilay(2013)}]{KB13}
Nate Kushman and Regina Barzilay. 2013.
\newblock Using semantic unification to generate regular expressions from
  natural language.
\newblock In \emph{Proceedings of the 2013 Conference of the North {A}merican
  Chapter of the Association for Computational Linguistics: Human Language
  Technologies (NAACL-HLT)}.

\bibitem[{Kwiatkowski et~al.(2013)Kwiatkowski, Choi, Artzi, and
  Zettlemoyer}]{kwiatkowski2013}
Tom Kwiatkowski, Eunsol Choi, Yoav Artzi, and Luke Zettlemoyer. 2013.
\newblock Scaling semantic parsers with on-the-fly ontology matching.
\newblock In \emph{Proceedings of the Conference on Empirical Methods in
  Natural Language Processing (EMNLP)}.

\bibitem[{Lee et~al.(2016)Lee, So, and Oh}]{lee}
Mina Lee, Sunbeom So, and Hakjoo Oh. 2016.
\newblock Synthesizing regular expressions from examples for introductory
  automata assignments.
\newblock In \emph{Proceedings of the 2016 ACM SIGPLAN International Conference
  on Generative Programming: Concepts and Experiences (GPCE)}.

\bibitem[{Lin et~al.(2018)Lin, Wang, Zettlemoyer, and Ernst}]{nl2bash}
Xi~Victoria Lin, Chenglong Wang, Luke Zettlemoyer, and Michael~D. Ernst. 2018.
\newblock {NL2Bash: A Corpus and Semantic Parser for Natural Language Interface
  to the Linux Operating System}.
\newblock In \emph{Proceedings of the International Conference on Language
  Resources and Evaluation {LREC}}.

\bibitem[{Ling et~al.(2016)Ling, Blunsom, Grefenstette, Hermann,
  Ko{\v{c}}isk{\'y}, Wang, and Senior}]{ling2016}
Wang Ling, Phil Blunsom, Edward Grefenstette, Karl~Moritz Hermann,
  Tom{\'a}{\v{s}} Ko{\v{c}}isk{\'y}, Fumin Wang, and Andrew Senior. 2016.
\newblock Latent predictor networks for code generation.
\newblock In \emph{Proceedings of the Annual Meeting of the Association for
  Computational Linguistics (ACL)}.

\bibitem[{Locascio et~al.(2016)Locascio, Narasimhan, DeLeon, Kushman, and
  Barzilay}]{deepregex}
Nicholas Locascio, Karthik Narasimhan, Eduardo DeLeon, Nate Kushman, and Regina
  Barzilay. 2016.
\newblock Neural generation of regular expressions from natural language with
  minimal domain knowledge.
\newblock In \emph{Proceedings of the Conference on Empirical Methods in
  Natural Language Processing (EMNLP)}.

\bibitem[{Luong et~al.(2015)Luong, Pham, and Manning}]{attention}
Thang Luong, Hieu Pham, and Christopher~D. Manning. 2015.
\newblock Effective approaches to attention-based neural machine translation.
\newblock In \emph{Proceedings of the Conference on Empirical Methods in
  Natural Language Processing (EMNLP)}.

\bibitem[{M\o{}ller(2017)}]{automaton}
Anders M\o{}ller. 2017.
\newblock dk.brics.automaton -- finite-state automata and regular expressions
  for {Java}.
\newblock \texttt{http://www.brics.dk/automaton/}.

\bibitem[{Nye et~al.(2019)Nye, Hewitt, Tenenbaum, and
  Solar-Lezama}]{sketchadapt}
Maxwell Nye, Luke Hewitt, Joshua Tenenbaum, and Armando Solar-Lezama. 2019.
\newblock Learning to infer program sketches.
\newblock In \emph{Proceedings of the International Conference on Machine
  Learning (ICML)}.

\bibitem[{Odena and Sutton(2020)}]{odena2020learning}
Augustus Odena and Charles Sutton. 2020.
\newblock Learning to represent programs with property signatures.
\newblock In \emph{Proceedings of the International Conference on Learning
  Representations (ICLR)}.

\bibitem[{Park et~al.(2019)Park, Ko, Cognetta, and Han}]{softregex}
Jun-U Park, Sang-Ki Ko, Marco Cognetta, and Yo-Sub Han. 2019.
\newblock {S}oft{R}egex: Generating regex from natural language descriptions
  using softened regex equivalence.
\newblock In \emph{Proceedings of the Conference on Empirical Methods in
  Natural Language Processing and the International Joint Conference on Natural
  Language Processing (EMNLP-IJCNLP)}.

\bibitem[{Paszke et~al.(2019)Paszke, Gross, Massa, Lerer, Bradbury, Chanan,
  Killeen, Lin, Gimelshein, Antiga, Desmaison, Kopf, Yang, DeVito, Raison,
  Tejani, Chilamkurthy, Steiner, Fang, Bai, and Chintala}]{pytorch}
Adam Paszke, Sam Gross, Francisco Massa, Adam Lerer, James Bradbury, Gregory
  Chanan, Trevor Killeen, Zeming Lin, Natalia Gimelshein, Luca Antiga, Alban
  Desmaison, Andreas Kopf, Edward Yang, Zachary DeVito, Martin Raison, Alykhan
  Tejani, Sasank Chilamkurthy, Benoit Steiner, Lu~Fang, Junjie Bai, and Soumith
  Chintala. 2019.
\newblock Pytorch: An imperative style, high-performance deep learning library.
\newblock In \emph{Advances in Neural Information Processing Systems
  (NeurIPS)}.

\bibitem[{Polosukhin and Skidanov(2018)}]{algolisp}
Illia Polosukhin and Alexander Skidanov. 2018.
\newblock Neural program search: Solving programming tasks from description and
  examples.
\newblock In \emph{Workshop at the International Conference on Learning
  Representations (ICLR Workshop)}.

\bibitem[{Quirk et~al.(2015)Quirk, Mooney, and Galley}]{ifttt}
Chris Quirk, Raymond Mooney, and Michel Galley. 2015.
\newblock Language to code: Learning semantic parsers for if-this-then-that
  recipes.
\newblock In \emph{Proceedings of the Annual Meeting of the Association for
  Computational Linguistics and the International Joint Conference on Natural
  Language Processing (EMNLP-IJCAI)}.

\bibitem[{Rabinovich et~al.(2017)Rabinovich, Stern, and Klein}]{Rabin17}
Maxim Rabinovich, Mitchell Stern, and Dan Klein. 2017.
\newblock Abstract syntax networks for code generation and semantic parsing.
\newblock In \emph{Proceedings of the Annual Meeting of the Association for
  Computational Linguistics (ACL)}.

\bibitem[{Ranta(1998)}]{Ranta:98}
Aarne Ranta. 1998.
\newblock A multilingual natural-language interface to regular expressions.
\newblock In \emph{Finite State Methods in Natural Language Processing}.

\bibitem[{Shin et~al.(2019)Shin, Allamanis, Brockschmidt, and
  Polozov}]{idiom2020}
Richard Shin, Miltiadis Allamanis, Marc Brockschmidt, and Oleksandr Polozov.
  2019.
\newblock Program synthesis and semantic parsing with learned code idioms.
\newblock In \emph{Advances in Neural Information Processing Systems
  (NeurIPS)}.

\bibitem[{Solar-Lezama(2008)}]{sketch}
Armando Solar-Lezama. 2008.
\newblock \emph{Program Synthesis by Sketching}.
\newblock Ph.D. thesis, University of California at Berkeley.

\bibitem[{Suhr et~al.(2018)Suhr, Iyer, and Artzi}]{suhr2018}
Alane Suhr, Srinivasan Iyer, and Yoav Artzi. 2018.
\newblock Learning to map context-dependent sentences to executable formal
  queries.
\newblock In \emph{Proceedings of the 2018 Conference of the North {A}merican
  Chapter of the Association for Computational Linguistics: Human Language
  Technologies (NAACL-HLT)}.

\bibitem[{Wang et~al.(2019)Wang, Titov, and Lapata}]{wang-2019-learning}
Bailin Wang, Ivan Titov, and Mirella Lapata. 2019.
\newblock Learning semantic parsers from denotations with latent structured
  alignments and abstract programs.
\newblock In \emph{Proceedings of the Conference on Empirical Methods in
  Natural Language Processing and the International Joint Conference on Natural
  Language Processing (EMNLP-IJCNLP)}.

\bibitem[{Wang et~al.(2016)Wang, Gulwani, and Singh}]{fidex}
Xinyu Wang, Sumit Gulwani, and Rishabh Singh. 2016.
\newblock {FIDEX: Filtering Spreadsheet Data Using Examples}.
\newblock In \emph{Proceedings of the 2016 ACM SIGPLAN International Conference
  on Object-Oriented Programming, Systems, Languages, and Applications
  (OOPSLA)}.

\bibitem[{Yaghmazadeh et~al.(2017)Yaghmazadeh, Wang, Dillig, and
  Dillig}]{sqlizer}
Navid Yaghmazadeh, Yuepeng Wang, Isil Dillig, and Thomas Dillig. 2017.
\newblock Sqlizer: query synthesis from natural language.
\newblock \emph{Proceedings of the ACM on Programming Languages},
  1(OOPSLA):1--26.

\bibitem[{Yin and Neubig(2017)}]{yin17}
Pengcheng Yin and Graham Neubig. 2017.
\newblock A syntactic neural model for general-purpose code generation.
\newblock In \emph{Proceedings of the Annual Meeting of the Association for
  Computational Linguistics (ACL)}.

\bibitem[{Zettlemoyer and Collins(2009)}]{zettlemoyer2009}
Luke Zettlemoyer and Michael Collins. 2009.
\newblock Learning context-dependent mappings from sentences to logical form.
\newblock In \emph{Proceedings of the Joint Conference of the Annual Meeting of
  the Association for Computational Linguistics {ACL}}.

\bibitem[{Zhong et~al.(2020)Zhong, Stern, and Klein}]{zhong2020}
Ruiqi Zhong, Mitchell Stern, and Dan Klein. 2020.
\newblock Semantic scaffolds for pseudocode-to-code generation.
\newblock In \emph{Proceedings of the Annual Meeting of the Association for
  Computational Linguistics (ACL)}.

\bibitem[{Zhong et~al.(2018{\natexlab{a}})Zhong, Guo, Yang, Peng, Xie, Lou,
  Liu, and Zhang}]{semregex}
Zexuan Zhong, Jiaqi Guo, Wei Yang, Jian Peng, Tao Xie, Jian-Guang Lou, Ting
  Liu, and Dongmei Zhang. 2018{\natexlab{a}}.
\newblock {S}em{R}egex: A semantics-based approach for generating regular
  expressions from natural language specifications.
\newblock In \emph{Proceedings of the Conference on Empirical Methods in
  Natural Language Processing (EMNLP)}.

\bibitem[{Zhong et~al.(2018{\natexlab{b}})Zhong, Guo, Yang, Xie, Lou, Liu, and
  Zhang}]{zexuan18}
Zexuan Zhong, Jiaqi Guo, Wei Yang, Tao Xie, Jian-Guang Lou, Ting Liu, and
  Dongmei Zhang. 2018{\natexlab{b}}.
\newblock Generating regular expressions from natural language specifications:
  Are we there yet?
\newblock In \emph{Workshops at the AAAI Conference on Artificial Intelligence
  (AAAI)}.

\end{thebibliography}

\end{document}